# Non-Destructive Peat Analysis using Hyperspectral Imaging and Machine Learning

Yijun Yan
*National Subsea Centre, Robert Gordon University, Aberdeen, U.K.*
*School of Computing, University of Dundee, Dundee, U.K.*
yijun.yan@ieee.org

Jinchang Ren*
*National Subsea Centre, Robert Gordon University, Aberdeen, U.K.*
j.ren@rgu.ac.uk
*Corresponding author

Barry Harrison
*The Scotch Whisky Research Institute, Research Avenue North, Edinburgh, UK.*
barry.harrison@swri.co.uk

Oliver Lewis
*The Scotch Whisky Research Institute, Research Avenue North, Edinburgh, UK.*
oliver.lewis@swri.co.uk

Yinhe Li
*National Subsea Centre, Robert Gordon University, Aberdeen, U.K.*
y.li24@rgu.ac.uk

Ping Ma
*National Subsea Centre, Robert Gordon University, Aberdeen, U.K.*
p.ma@rgu.ac.uk

*Abstract*—Peat, a crucial component in whisky production, imparts distinctive and irreplaceable flavours to the final product. However, the extraction of peat disrupts ancient ecosystems and releases significant amounts of carbon, contributing to climate change. This paper aims to address this issue by conducting a feasibility study on enhancing peat use efficiency in whisky manufacturing through non-destructive analysis using hyperspectral imaging. Results show that shot-wave infrared (SWIR) data is more effective for analyzing peat samples and predicting total phenol levels, with accuracies up to 99.81%.

## I. INTRODUCTION

As environmental concerns grow, the whisky industry faces increasing pressure to reduce its environmental footprint, notably in relation to peat usage [1]. The unique flavours contributed by peat have led distilleries to seek ways to optimize its use in response to the global demand for peated whiskies [2]. By gaining a deeper understanding of the characteristics of various peat sources, distilleries can refine their production processes, ensuring distinct and consistent flavours while improving efficiency in peat usage. This not only presents potential cost savings but also aligns with the urgent need for environmental responsibility.

The main technological challenge is to develop and apply advanced techniques that can accurately quantify the phenolic levels in peat without damaging the sample and with minimal time, financial cost, and environmental impact. Addressing this challenge opens up a market opportunity to improve quality control, optimize whisky production processes, and enhance the overall competitiveness of whisky manufacturers. Existing methods for peat analysis, such as High Performance Liquid Chromatography (HPLC) or Gas Chromatography (GC), involve extensive preparation, potentially leading to material waste [3]. These methods also use expensive solvents, which increase the overall cost of analysis and require careful disposal to minimize environmental harm. Additionally, maintenance and operation costs are high due to the need for specialized expertise and equipment. Currently, there is no direct means to measure phenols in peat. To tackle this challenge, this project pioneers the application of hyperspectral imaging for peat analysis, a novel endeavour that has not been previously explored.

Hyperspectral Imaging (HSI) is a technique that combines spectroscopy with digital imaging. Regular multispectral systems, such as RGB cameras, collect information in a limited number of distinct wavebands spread out over a certain spectral range. HSI in contrast captures intensities over a continuous spectral range in very narrow wavebands. Each pixel does not only represent spatial information in the form of x and y coordinates but also spectral information in form of a continuous spectrum. Depending on the system, this can entail several hundred wavebands. The data is stored in a three-dimensional data cube, often referred to as a hypercube where for each wavelength, a full resolution spatial image is available. Due to recent advances in imaging technology in the last decades, HSI has been widely used in various food and agricultural scenarios, such as meat quality assessment [4], tea quality evaluation [5], and pineapple grading [6]. This paper aims to extend the utility of these proven hyperspectral imaging technologies to the new and promising realm of peat analysis.

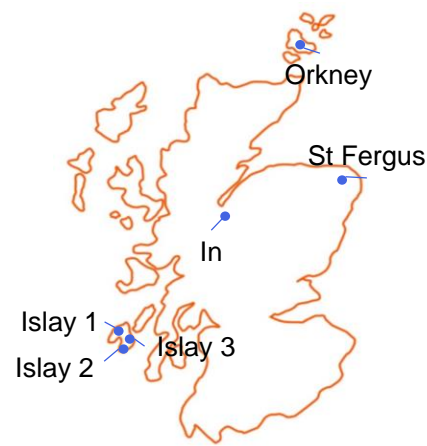

Fig. 1. Sources for peat samples

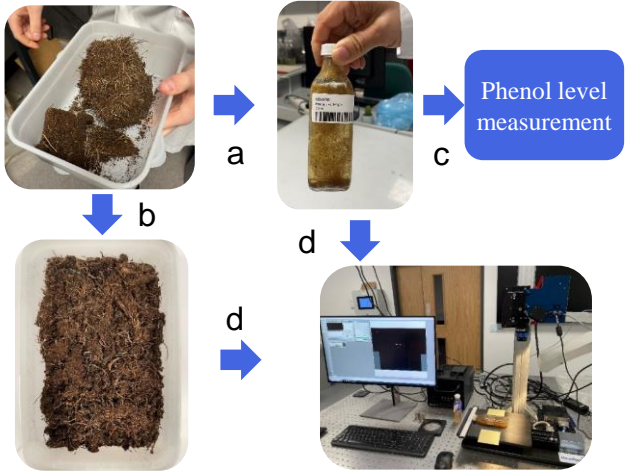

Fig. 2. Sample preparation, (a) condensation process, (b) smashing and storage, (c) HPLC, (d) data scanning.

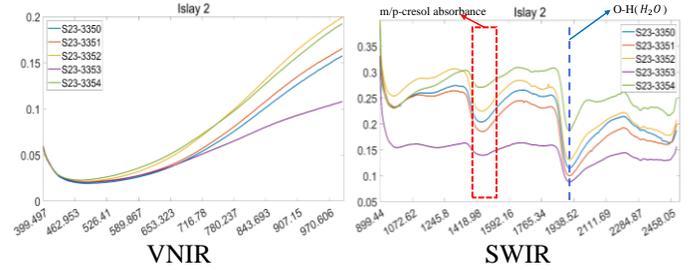

Fig. 3. The mean spectra of a number of random pixels selected from 5 Islay 2 peat samples.

## II. MATERIAL

35 categories of peat samples are used in our experiment. These peat samples were from six sources: St Fergus, In, Islay 1, Islay 2, Orkney and Islay 3 (Fig. 1). Peat samples were randomly sampled from peat stacks ready to be used in the peating process. For each peat sample, condensation process was carried out to generate 35 condensates. HPLC was used to measure phenols levels in peat smoke condensates. Other properties such as organic matter was determined by loss on ignition, and moisture content was determined by oven drying method. For the purposes of HSI data acquisition, each peat sample was smashed and stored in a rectangle container (Fig. 2) and the corresponding condensate was stored in glass bottle.

## III. METHODOLOGY

### A. Data acquisition

Both visible-near infrared camera (VNIR) and shortwave-infrared (SWIR) camera were used for data acquisition. All cameras operate in pushbroom mode, meaning that the camera is pointed downwards and scans only a single line at a time. The objects to be scanned (i.e., peat container and condensate bottle) are moved with a translational stage at even speed underneath the camera and are thereby fully scanned. This means that the only spatial limitation is the width of the objects. They can be theoretically infinitely long, only limited by the storage capacity. Technical details of both systems and the camera settings in the lab environment are shown in Table 1.

### B. Data processing

The acquired raw spectrum data, denoted as s, from each camera underwent radiometric calibration and was subsequently converted to reflectance, represented as r, using the Eq.(1):

$$r = \frac{s - d}{w - d} \quad (1)$$

where d is the dark reference spectrum, obtained by imaging without exposing the camera to the light, w represents the white reference spectrum acquired by imaging an ideally reflective white surface, such as Spectralon, which exhibits Lambertian scattering properties.

TABLE 1 DETAILS OF IMAGING SYSTEM

|  | VNIR camera | SWIR camera | HSI scanning system |
|---|---|---|---|
| Wavelength range (nm) | 400-1000 nm | 900-2500 nm | |
| Spectral bands | 371 bands | 270 bands | |
| Spatial bands | 1600 pixels | 640 pixels | |
| Working distance (cm) | 40 | 40 | |
| Speed of translation stage (mm/s) | 2.4 | 9.2 | |

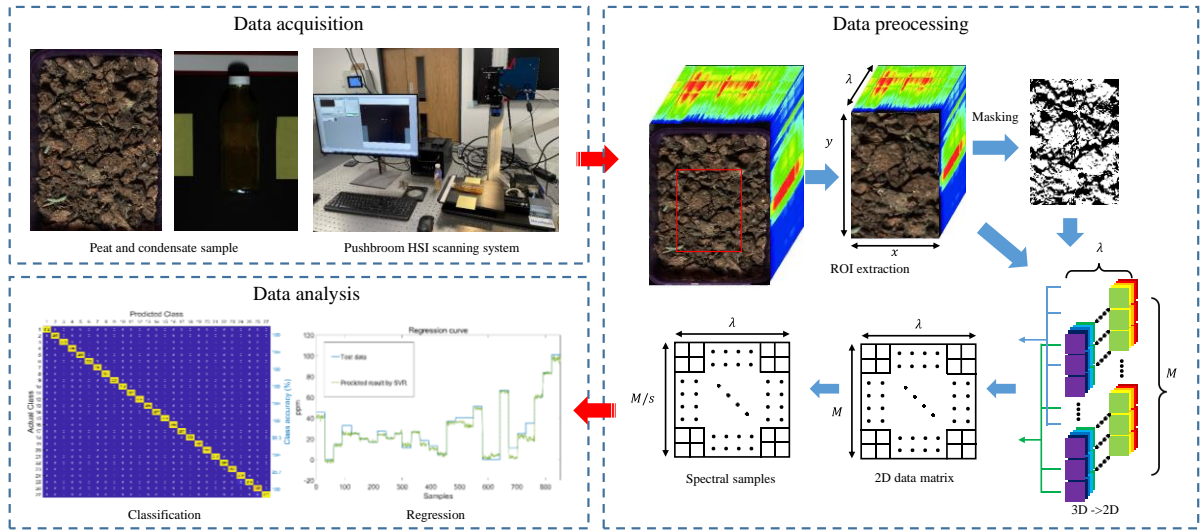

Fig. 4. Data processing workflow

Taking the Islay 2 peat as examples, the typical spectral of the two cameras obtained after radiometric calibration are depicted in Fig. 3. The VNIR profile of different peat sample have very similar trend and contains hardly any useful information but intensity differences. The SWIR camera is able to pick up spectral information such as O-H absorption band and m/p-cresol absorption bands.

Fig. 4 illustrates the proposed workflow of data processing. Following data acquisition, the region of interest (ROI) in the central area is extracted to eliminate regions near the boundaries that contain shadows. Subsequently, an adaptive thresholding technique [7] is applied to the ROI to further remove shadow regions caused by an uneven surface that may obstruct the light from being evenly distributed on the peat. Let M denote the number of pixels in masked region of interest, and we randomly selected s pixels and calculate their average value as a representative spectral sample to compensate for the fact that the distribution of phenols on the surface is very uneven. To this end, we would have M/s spectral samples extracted from the hypercube. For SWIR data, the average number of pixels for each class of peat and condensate is $1.04*10^5$ and $1.44*10^4$ respectively. For VNIR data, the average number of pixels for each class of peat and condensate is $4*10^5$ and $9*10^4$ respectively. When s=50, the number of spectral sample for SWIR peat data, SWIR condensate data, VNIR peat data and VNIR condensate data is 2080, 288, 8000, and 1800, respectively.

In the experiment, we assess the feasibility of our approach by performing two tasks: grading and property estimation. The grading task is treated as a classification problem, where each peat sample is considered an individual category with a specific quality grade. The property estimation task is viewed as a regression problem, where the objective is to predict properties such as total phenol level, moisture, and organic matter content in the peat.

To address these tasks, we employ a widely used classification model, Support Vector Machine (SVM). SVM is a powerful machine learning algorithm that can effectively handle both classification and regression problems. To ensure optimal performance, we utilize a grid search technique [8] to determine the best hyperparameters for the SVM model.

## IV. RESULTS AND DISCUSSION

Table 2 and Table 3 investigate how the selection of the parameter s affected the grading results using SWIR and VNIR data, respectively. OA, AA, and KP indicate the overall accuracy, average accuracy, and Cohen's Kappa, which aim to measure the degree of agreement between the classifier and the known concentration. 5% spectral samples were used for training and the rest spectral samples were used for testing. When s =50, it can be observed that SWIR data is significantly more useful for analyzing peat samples, with the overall accuracy for grading reaching up to 99.65%. On the other hand, VNIR data performs better on condensate samples, where the OA can reach up to 99.33%. When s is reduced to 10, SWIR data outperforms VNIR data on both peat and condensate data. The main reason for this is the presence of more meaningful information in SWIR data compared to VNIR data. Furthermore, if promising results can be achieved with a low value of s, point-scanning hyperspectral sensors can potentially be used to reduce costs while maintaining grading accuracy.

TABLE 2 CLASSIFICATION ACCURACIES FOR DIFFERENT SELECTION OF S IN SWIR DATA (BEST RESULTS ARE HIGHLIGHTED IN BOLD)

| | Condensate samples | | | | | Peat samples | | | | |
|---|---|---|---|---|---|---|---|---|---|---|
| s | 10 | 20 | 30 | 40 | 50 | 10 | 20 | 30 | 40 | 50 |
| OA | 94.91 | 97.33 | 97.69 | 98.06 | 98.37 | 96.28 | 98.58 | 99.27 | 99.58 | **99.65** |
| AA | 94.91 | 97.33 | 97.69 | 98.06 | 98.37 | 96.33 | 98.60 | 99.28 | 99.59 | **99.66** |
| KP | 94.74 | 97.24 | 97.62 | 98.00 | 98.32 | 96.17 | 98.53 | 99.25 | 99.57 | **99.64** |

TABLE 3 CLASSIFICATION ACCURACIES FOR DIFFERENT SELECTION OF S IN VNIR DATA (BEST RESULTS ARE HIGHLIGHTED IN BOLD)

| | Condensate samples | | | | | Peat samples | | | | |
|---|---|---|---|---|---|---|---|---|---|---|
| s | 10 | 20 | 30 | 40 | 50 | 10 | 20 | 30 | 40 | 50 |
| OA | 92.96 | 97.27 | 98.50 | 99.06 | **99.33** | 69.47 | 81.01 | 86.48 | 89.71 | 91.78 |
| AA | 92.96 | 97.27 | 98.50 | 99.06 | **99.31** | 69.14 | 80.76 | 86.31 | 89.57 | 91.68 |
| KP | 92.72 | 97.18 | 98.45 | 99.03 | **99.33** | 68.55 | 80.45 | 86.08 | 89.40 | 91.53 |

TABLE 4 TOTAL PHENOL LEVEL PREDICTION ON SWIR AND VNIR DATA WITH s = 50.

|  | SWIR | | VNIR | |
| --- | --- | --- | --- | --- |
|  | Condensate samples | Peat samples | Condensate samples | Peat samples |
| MAE (ppm) | 8.49 | **3.28** | 7.38 | 17.67 |
| RMSE (ppm) | 11.66 | **4.73** | 10.59 | 26.03 |
| $R^2$ (%) | 95.21 | **99.25** | 96.05 | 75.08 |

TABLE 5 MOISTURE AND ORGANIC MATTER PREDICTION WITH s = 50

| Moisture | Condensate samples | Peat samples | OM | Condensate samples | Peat samples |
| --- | --- | --- | --- | --- | --- |
| MAE (%) | 3.53 | 0.580 | MAE (%) | 1.29 | 0.534 |
| RMSE (%) | 5.20 | 0.778 | RMSE (%) | 1.81 | 0.776 |
| $R^2$ (%) | 94.68 | 99.88 | $R^2$ (%) | 94.92 | 99.03 |

Table 4 presents the total phenol level prediction results on SWIR and VNIR data with s = 50. Once again, SWIR data produces the best results, with mean absolute error (MAE), root mean square error (RMSE) and $R^2$ reaching 3.28ppm, 4.73ppm, and 99.25%, respectively.

According to the findings from Table 2-4, for peat samples, SWIR data demonstrated superior performance in grading accuracy and total phenol level prediction compared to VNIR data. Consequently, we chose to use SWIR data to assess the effectiveness of moisture and organic matter (OM) prediction in peat. Similarly, for condensate samples, VNIR data exhibited better grading accuracy than SWIR data, making it the preferred choice for evaluating moisture and OM prediction in condensate. The results shown in Table 5 confirm the excellent performance of the selected data types for each sample category. The high accuracy obtained in predicting moisture and OM content using SWIR data for peat and VNIR data for condensate underscore the effectiveness of our approach.

## V. CONCLUSION

This study has successfully demonstrated the immense potential of Hyperspectral Imaging (HSI) for assessing peat quality and accurately measuring phenolic content in both peat and liquid samples. The first study condensed 50 pixels into a single spectral sample and utilized only 5% of these samples for training. The model achieved impressive grading overall accuracies of 99.65% for peat and 99.33% for condensate samples. The prediction of total phenol level, moisture and organic matter can reach up to $R^2$ of 99.25%, 99.88%, and 99.03%, respectively. These results underscore the capability of HSI to provide accurate assessments even with limited training data.

Throughout our study, SWIR data proved more effective than VNIR data for phenolic measurements on peat sample and the best grading accuracies, while VNIR data performs better than SWIR data on condensate samples. Overall, our findings establish HSI as a powerful tool for non-destructive, rapid, and accurate assessment of peat quality and phenolic content.

ACKNOWLEDGMENT

We greatly appreciate the funding support from Innovate UK, and the Scotch Whisky Research Institute who provided the peat samples in the experiments.